%% file: acl2023.tex
\pdfoutput=1

\documentclass[11pt]{article}

\usepackage{ACL2023}

\usepackage{times}
\usepackage{latexsym}
\usepackage{tabularx}

\usepackage[T1]{fontenc}

\usepackage[utf8]{inputenc}

\usepackage{microtype}

\usepackage{inconsolata}

\usepackage{graphicx}
\usepackage{multirow}
\usepackage{booktabs}
\usepackage{hhline}

\usepackage{longtable}

\newif\ifcomments
\ifcomments
\newcommand{\draftcomment}[3]{{\color{#2}[\textsc{#1} #3]}}
\else
\newcommand{\draftcomment}[3]{}
\fi

\newcommand{\di}[1]{\draftcomment{Di:}{blue}{#1}}

\newcommand{\joel}[1]{\draftcomment{Joel:}{orange}{#1}}

\title{Event Extraction as Question Generation and Answering}

\author{
    Di Lu \quad
    Shihao Ran \quad 
    Joel Tetreault \quad 
    Alejandro Jaimes \\
    Dataminr Inc. \\
    \texttt{\{dlu,sran,jtetreault,ajaimes\}@dataminr.com}
}

\begin{document}
\maketitle
\input{0_abstract}
\input{1_intro}
\input{2_model}
\input{3_experiment}
\input{4_results}
\input{5_conclusion}
\input{6_limitations}

\input{8_Acknowledgements}



\bibliography{anthology,custom}
\bibliographystyle{acl_natbib}

\appendix
\input{appex_b}

\end{document}

%% file: 0_abstract.tex
\begin{abstract}

Recent work on Event Extraction has reframed the task as  Question Answering (QA), with promising results.\joel{Di - is it safe to say that most EE methods use QA and these are the most performant?}\di{it might not be safe to claim that. some other generation-based methods and some graph-based method have comparable performance.}
The advantage of this approach is that it addresses the error propagation issue found in traditional token-based classification approaches by directly predicting event arguments without extracting candidates first. However, the questions are typically based on fixed templates and they rarely leverage contextual information such as relevant arguments. In addition, prior QA-based approaches have difficulty handling cases where there are multiple arguments for the same role. In this paper, we propose QGA-EE, which enables a Question Generation (QG) model to generate questions that incorporate rich contextual information instead of using fixed templates. We also propose dynamic templates to assist the training of QG model.
Experiments show that QGA-EE outperforms all prior single-task-based models on the ACE05 English dataset.\footnote{Our code is available at \url{https://github.com/dataminr-ai/Event-Extraction-as-Question-Generation-and-Answering} for research purposes.}


\end{abstract}


%% file: 1_intro.tex
\section{Introduction}
Event Extraction (EE) aims to extract core information elements (e.g. who, what, where, when) from text, and is a very important task in Natural Language Processing (NLP). It provides inputs to downstream applications such as Summarization~\cite{filatova-hatzivassiloglou-2004-event}, Knowledge Base Population~\cite{ji2011knowledge}, and Recommendation~\cite{lu2016cross}. 

\begin{figure}[ht]
    \centering
    \includegraphics[width=0.9\linewidth]{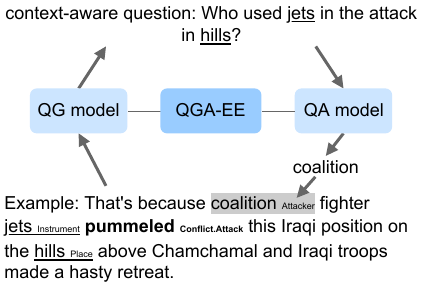}
    \caption{An event mention example from ACE. An ACE \texttt{Conflict.Attack} event with \textit{pummeled} as trigger word and three event arguments: \textit{coalition} (\texttt{Attacker}),  \textit{jets} (\texttt{Instrument}) and \textit{hills} (\texttt{Place}).}
    \label{fig:ace_intro}
\end{figure}

Previous work~\cite{li2013joint,nguyen2016joint,sha2018jointly} is typically based on a pipeline approach, which first identifies the event trigger word/phrase and argument candidates, and then applies a classifier to the pair-wise features to classify the roles of the candidates.  Unfortunately, errors tend to propagate down the pipeline.
Recently, some approaches have formulated EE as a Question Answering (QA) problem~\cite{du2020event,li2020event,lyu2021zero}\joel{you mention some approaches but then just cite one paper.  good idea to list more}\di{added more EE-QA paper} to mitigate the issue, in which questions for each argument role are manually defined by templates. 
For example, extracting the \texttt{Attack} argument from the \texttt{Conflict.Attack} event in the sentence \textit{`That's because coalition fighter jets pummeled this Iraqi position on the hills  above Chamchamal and Iraqi troops made a hasty retreat.'} is reframed as answering the question \textit{`Who was the attacking agent?'}\joel{what does it mean to convert to a task?  it sounds weird}\di{rewrote it} 
These approaches have shown promising results, but template-based questions are limiting: since the templates are built manually, they are fixed and rarely include contextual information (i.e., specific to the inputs), except for trigger words in some work~\cite{du2020event}. Formulating good questions, however, has been shown to improve performance for standard QA tasks~\cite{rajpurkar2018know}. 
For QA-based EE, a question that incorporates richer contextual information such as other event arguments could yield better results (e.g.  \textit{`Who used \underline{jets} in the attack in \underline{hills}?'} in Figure~\ref{fig:ace_intro}). 
\joel{do you mean to say "could produce better"?} 

In this paper, we propose QGA-EE, which consists of 1) a QG model for generating a context-aware question conditioned on a target argument role and 2) a QA model for answering the context-aware question to extract the event argument. We also design dynamic templates to generate the gold context-aware questions for QG model training.

To the best of our knowledge, this is the first QA-based EE work that utilizes dynamic templates and focuses on generating context-aware questions.
\citet{li2020event} also propose a model to generate questions that incorporate contextual information for both event trigger and arguments. However, our work has two main advantages. First, in \citet{li2020event} the question only incorporates the contextual information at the ontology level (e.g. argument role, event type). In our work, the generated questions incorporate contextual information at an event mention-level. For example, the question generated by our model includes the real event argument rather than just the argument role (e.g. ‘hills’ vs ‘Place’). Second, the questions in their work are generated by filling in the templates, but our templates are dynamic and used to train the QG model which can automatically generate the optimal question given a specific event mention and the concerned argument role.

Experimental results show that QGA-EE outperforms all of the single-task-based models on the Automatic Content Extraction (ACE) 2005 English dataset~\cite{doddington2004automatic} and even achieves competitive performance with state-of-the-art joint IE models.

%% file: 2_model.tex
\section{Model}
Figure~\ref{fig:ace_intro} shows the overall framework of QGA-EE.
It focuses on Event Argument Extraction (EAE) only, but can be paired with any event trigger tagger to perform end-to-end EE. In Section~\ref{sec:results}, we pair it with a standard sequence labeling trigger tagger to evaluate its end-to-end EE performance. 








\input{2.2_qg_model}
\input{2.3_qa_model}

%% file: 2.2_qg_model.tex
\subsection{Question Generation Model}


Previous QA-based EE work~\cite{du2020event} fills in pre-designed templates with trigger information to generate the input questions to the QA model. However missing contextual information in the questions is a bottleneck for the performance of the QA model.

QGA-EE uses a QG model to generate context-aware questions conditioned on the input sentence and target role, which is based on a sequence-to-sequence architecture (e.g. BART\cite{lewis2019bart}, T5\cite{roberts2019exploring}). In order to train the QG model, we design \textbf{Dynamic Templates} for each role in the ACE ontology.\footnote{\url{https://www.ldc.upenn.edu/sites/www.ldc.upenn.edu/files/english-events-guidelines-v5.4.3.pdf}} We design multiple templates for each role, and each of them includes different combinations of other argument roles.

\begin{table}[th]
    \centering
    \small
    \begin{tabular}{l}
    \hline
        Who was the attacking agent? \\
        Who attacked [\texttt{Target}]?\\
        Who used [\texttt{Instrument}] in the attack?\\
        Who made the attack in [\texttt{Place}]?\\
        Who attacked [\texttt{Target}] using [\texttt{Instrument}]?\\
        Who attacked [\texttt{Target}] in [\texttt{Place}]?\\
        Who used [\texttt{Instrument}] in the attack in [\texttt{Place}]?\\
        Who attacked [\texttt{Target}] using [\texttt{Instrument}] in [\texttt{Place}]? \\ 
        \hline
    \end{tabular}
    \caption{Dynamic templates for \texttt{Attacker} role in \texttt{Conflict.Attack} event with different combinations of known argument roles based on ACE ontology.}
    \label{tab:templates}
\end{table}


For example, the \texttt{Conflict.Attack} event in ACE has four predefined argument roles: \texttt{Attacker}, \texttt{Target}, \texttt{Instrument} and \texttt{Place}.\footnote{We follow the experimental setting of prior work, which excludes all the \texttt{Value} and \texttt{Timex}. Thus the argument roles such as \texttt{Time} are not included.}
For the \texttt{Attacker} role, we exhaustively design eight templates using all of the possible combinations of the other roles included in the question (Table~\ref{tab:templates}). 
When the model fills in the templates given a specific event mention, it is common that some of the predefined argument roles do not exist in the event mention. Thus the model only keeps the templates that contain the slots for argument roles appearing in the event mention.
For the example in Figure~\ref{fig:ace_intro}, the \texttt{Target} role is not mentioned. So we ignore all of the templates that contain the \texttt{[Target]} slot, and we obtain four candidate questions for the \texttt{Attacker} role with corresponding arguments filled in: (1)\textit{Who was the attacking agent?} (2) \textit{Who used jets in the attack?} (3) \textit{Who made the attack in hills?} (4) \textit{Who used jets in the attack in hills?}

To train a QG model to generate the questions that cover as many contextual information as possible, we use the question that contains the most contextual arguments as the ground truth. For the example in Figure~\ref{fig:ace_intro}, we choose the question \textit{`Who used jets in the attack in hills?'}, because it contains two arguments: \textit{`jets'} and \textit{`hills'}, the other three candidate questions listed above contain one or zero arguments. If more than one candidate question contains the most contextual arguments, we then pick the first one. The input and output examples for the QG model are as follows: 

\noindent\fbox{
 \parbox{\linewidth}{
 \textbf{Input:} role: attacker context: That's because coalition fighter jets * pummeled * this Iraqi position on the hills above Chamchamal and Iraqi troops made a hasty retreat.
 
 \textbf{Output:} Who used jets in the attack in hills?
}
 }

%% file: 2.3_qa_model.tex
\subsection{Question Answering Model}
\label{sec:qa_model}
Different from prior QA-based EE work that adapt an encoder-only architecture and predict the offsets of the event arguments~\cite{chen2019reading, du2020event,li2020event,liu2020event,feng2020probing,lyu2021zero,zhou2021role}, our QA model is based on a sequence-to-sequence architecture (e.g. BART, T5), and generates the answer string directly.
This enables prediction of multiple event arguments that are associated with the same role. \citet{li2021document} also adapts a generation model, but the input template is fixed. The examples of input and output are as follows:





\noindent\fbox{
 \parbox{\linewidth}{
 \textbf{Input:} \textit{question: Who was harmed in * injured * event? context: \underline{Injured} Russian diplomats and a convoy of America's Kurdish comrades in arms were among unintended victims caught in crossfire and friendly fire Sunday.}
 
 \textbf{Output:} \textit{diplomats; convoy; victims $<$ /s $>$}
}
 }

\noindent\textbf{Post-processing}
We split the output into a list of candidates (by \texttt{`;'}), and retrieve the arguments with offsets by exactly matching against the original sentence.
We dynamically change the start position for searching to preserve the order of the retrieved event arguments. 
If an argument candidate cannot be matched with the original sentence, we discard it.
Unlike the QG model, we use all of the possible questions as inputs during training for data augmentation purposes, and the size of the training data increases from 15,426 to 20,681.\joel{what do those numbers refer to?}\di{rewrote}
But in the testing phase, we use the single question generated by the QG model for each argument role.

%% file: 3_experiment.tex
\section{Experimental Setup}
\input{3.1_data.tex}
\input{3.2_baseline}
\input{3.3_implementation.tex}

%% file: 3.1_data.tex
\subsection{Dataset and Evaluation Metrics}

We conduct the experiments on the ACE 2005 English corpora,
which has 33 event types and 22 argument roles. It contains 599 documents collected from newswire, weblogs, broadcast conversations, and broadcast news.
More specifically, we follow the pre-processing steps in \citet{wadden2019entity},\footnote{\url{https://github.com/dwadden/dygiepp}} and evaluate our models on the resulting ACE05-E dataset. 

 For evaluation, we use the same criteria as prior work~\cite{li2013joint}: An \textbf{event trigger} is correctly identified if its offsets exactly match a reference. It is correctly classified if both its offsets and event type match a reference. An \textbf{event argument} is correctly identified (Arg-I) if its offsets and event type match a reference in the ground truth. It is correctly classified (Arg-C) if all of its offsets, event type, and argument role match a reference.
 

%% file: 3.2_baseline.tex
\subsection{Compared Baselines}
\noindent\textbf{Model Variants.} To evaluate the generalizability of our approach, we evaluate two QGA-EE variants: \textbf{QGA-EE$_{BART}$} and \textbf{QGA-EE$_{T5}$}, which use BART and T5 as backbones respectively.

We compare the proposed models against SOTA EE models. \textbf{BERT QA}~\cite{du2020event} use BERT as the encoder and predict the positions of the argument directly with role-driven questions.
\textbf{TANL}~\cite{paolini2021structured} transfers input sentences into augmented natural language sentences for structured prediction. \textbf{TEXT2EVENT}~\cite{lu2021text2event} is a sequence-to-structure network for event extraction.\footnote{DEGREE~\cite{hsudegree} is not included because it is not evaluated on all of the argument roles.} \citet{ma2020resource} utilizes dependency parses as additional features. \textbf{BART-Gen}~\cite{li2021document} is a BART-based generation model proposed for document-level event extraction.

We also compare with joint IE models trained on all of the ACE annotations which include entities, relations, and events. They benefit from additional information from other tasks and usually achieve better performance than the models trained on a single task. 
It is not fair to directly compare our model with the joint models since they incorporate more information beyond the standard EE training sets, but we still list their scores as a reference. \textbf{DYGIE++}~\cite{wadden2019entity} is a BERT-based model that models span representations with within-sentence and cross-sentence context. \textbf{ONEIE}~\cite{lin2020joint} leverages global features. \textbf{FourIE}~\cite{van2021cross} and \textbf{GraphIE}~\cite{van2022joint} are Graph Convolutional Networks-based models and \textbf{AMR-IE}~\cite{zhang2021abstract} utilizes AMR~\cite{banarescu2013abstract} parser.


%% file: 3.3_implementation.tex
\subsection{Implementation Details}
We conduct all of the experiments on a single V100 GPU. For finetuning, we use the Adafactor~\cite{shazeer2018adafactor} optimizer with a learning rate of $1*10^{-4}$, weight decay of $1 * 10^{-5}$,  and clip threshold of $1.0$. We train the model for 20 epochs. Further details such as hyperparameters and data statics for model training and evaluation are in Appendix~\ref{app:implement}.

%% file: 4_results.tex
\section{Results}
\label{sec:results}
\input{4.1_overall}
\input{4.2_ablation}
\input{4.3_data_aug}
\input{4.4_analysis}

%% file: 4.1_overall.tex
\subsection{Event Argument Extraction Performance}
\begin{table}[!th]
    \centering
    \small
    \begin{tabular}{l|c|c}
    \hline
    {} & {Arg-I} & {Arg-C}\\
    \hline

         {BERT\_QA~\cite{du2020event}}&{68.2}&{65.4}  \\
         {TANL$^{+}$~\cite{paolini2021structured}}& {65.9} & {61.0}\\
        
         {\citet{ma2020resource}}&{-} & {62.1}\\
         {BART-Gen~\cite{li2021document}}& {69.9} & {66.7}\\
    \hline
    
        {DYGIE++$^{*+}$~\cite{wadden2019entity}}&{66.2}&{60.7}  \\
         {ONEIE$^{*+}$~\cite{lin2020joint}} & {73.2} & {69.3}\\
         \hline
        {QGA-EE$_{BART}$ (ours)} &  {72.4} & {70.3} \\
        {QGA-EE$_{T5}$ (ours)} & {\textbf{75.0}} & {\textbf{72.8}}  \\
    \hline
    \end{tabular}
    \caption{Event Extraction Results on ACE05-E test data (F1, \%) with gold triggers. $^{*}$ models are trained with additional entity and relation data. $^{+}$ numbers are reported from~\citet{hsudegree}, and others are from the original papers.}
    \label{tab:arg_overall_gold}
\end{table}
Table~\ref{tab:arg_overall_gold} shows the performance of QGA-EE models on ACE05-E test set with gold triggers.\footnote{Performance of FourIE, AMR-IE and GraphIE in gold triggers are not available in their original papers.}
Both QGA-EE variants outperform all other approaches, and using T5 as backbone provides an improvement of 2.5\% over BART.
The improvement over the prior QA-based models BERT\_QA shows that generation-based QA models are more effective than position-based QA models for EE.
QGA-EE$_{BART}$ outperforms the BART-based baseline BART-Gen and QGA-EE$_{T5}$ outperforms the T5-based baseline TANL, which demonstrates the effectiveness of our models with different backbones.
Our models even outperform the joint IE models DYGIE++ and ONEIE, which leverage additional information from entities and relations.

\subsection{Event Extraction Performance}

\begin{table}[!th]
    \centering
    \small
    \begin{tabular}{l|c|c}
    \hline
    {} & {Arg-I} & {Arg-C}\\
    \hline
         {BERT\_QA~\cite{du2020event}}&{54.1}&{53.1}  \\
         {TANL~\cite{paolini2021structured}}& {50.1} & {47.6}\\
         {TEXT2EVENT~\cite{lu2021text2event}}& {-} & {53.8}\\
         {\citet{ma2020resource}}& {56.7} & {54.3}\\
         {BART-Gen~\cite{li2021document}}& {-} & {53.7}\\
    \hline
         {DYGIE++$^{*}$~\cite{wadden2019entity}}&{54.1}&{51.4}  \\
         {ONEIE$^{*}$~\cite{lin2020joint}} & {59.2} & {56.8}\\
         {FourIE$^{*}$~\cite{van2021cross}} & {{60.7}} & {{58.0}}\\
         {AMR-IE$^{*}$~\cite{zhang2021abstract}} & {\textbf{60.9}} & {{58.6}}\\
         {GraphIE$^{*}$~\cite{van2022joint}} & {\textbf{-}} & {\textbf{59.4}}\\
         
         \hline
        {QGA-EE$_{BART}$ (ours)} & {57.1}& {55.6} \\
        {QGA-EE$_{T5}$ (ours)} & {59.8} & {57.9} \\
        

    \hline
    \end{tabular}
    \caption{Event Extraction Results on ACE05-E test data (F1, \%) with predicted triggers. $^{*}$ models are trained with additional entity and relation data. All numbers of baselines are reported from the original papers.}
    \label{tab:arg_overall}
\end{table}
We also evaluate our models on ACE05-E  in a more ``real world'' fashion with \textit{predicted} triggers extracted by an ALBERT-based~\cite{lan2019albert} sequence-labeling model (Table~\ref{tab:arg_overall}).\footnote{The model is trained on ACE05-E and the F1 score on test set is 72.96\%. More details in Appendix.}
Similar to the performance on gold triggers, QGA-EE benefits more from the T5 backbone on predicted triggers. Both QGA-EE variants outperform all the EE-task-centered baselines by more than 1\% on Arg-C. 

\joel{also Tables 2 and 3 should delineate more clearly which ones are single task, and which are joint, otherwise we're going to run into the same issue as in AAAI}\di{added notes in captions}
We also include the scores from SOTA joint IE models, DYGIE++, ONEIE, FourIE, AMR-IE and GraphIE, as reference. But, as stated earlier, it is not fair to compare our models directly with them, as they benefit from being trained with all of the annotations from entities, relations, and events. 
Also it should be noted that their trigger labeling models have more complicated architectures and thus perform  significantly better than the sequence-labeling based tagger we use (F1 75.4\% from FourIE and F1 74.7\% from OneIE). This further boosts the end-to-end EE performance. 

%% file: 4.2_ablation.tex
\subsection{Ablation Study}

Table~\ref{tab:abalation_gold} shows the ablation study of the QGA-EE$_{T5}$ model on the ACE05 test set with gold triggers. By replacing the QG model with simple context-unaware templates, the F1 score decreases by $1.65\%$. It demonstrates that the context-aware questions generated by our QG component enhance the end-to-end event argument extraction performance. Additionally, the generation-based QA model deals with multi-argument situations better and provides an improvement of  $4.24\%$.

\begin{table}[ht]
    \centering
    \small
    \begin{tabular}{l|cc}
    \hline
    {} & {Arg-I} & {Arg-C}\\
    \hline
        {QGA-EE$_{T5}$} & {75.04} & {72.78}  \\
    \hline
        
        {- w/o pretrained QG} & {73.57} & {71.13} \\
    \hline
        {- w/o pretrained QG \& mutli-arg support} & {69.61} & {66.89} \\
    \hline
        
    \end{tabular}
    \caption{Ablation study with gold triggers on ACE05-E test set (F1, \%).}
    \label{tab:abalation_gold}
\end{table}

        



%% file: 4.3_data_aug.tex
\subsection{Impact of Data Augmentation}
As we mentioned in Section~\ref{sec:qa_model}, the size of the training data increases from 15,426 to 20,681 as a benefit of our proposed dynamic templates. To evaluate the contribution of the data augmentation, we evaluate the performance of QGA-EE on ACE05 test data with partial training data (with gold triggers). With 40\% of the training examples after data augmentation (8,272), QGA-EE achieves a F1 score of 71.42\% on ACE05-E test set with gold triggers. It outperforms all of the baselines in Table~\ref{tab:arg_overall_gold}, which demonstrates the effectiveness of our proposed model.

\begin{table}[ht]
    \centering
    \small
    \begin{tabular}{l|cc}
    \hline
    {} & {Arg-I} & {Arg-C}\\
    \hline
        {QGA-EE$_{T5}$ with 100\% training data} & {75.04} & {72.78}  \\
    \hline
        
        {QGA-EE$_{T5}$ with 80\% training data} & {73.86 } & {71.64} \\
    \hline
        {QGA-EE$_{T5}$ with 60\% training data} & { 73.15 } & {71.63 } \\
    \hline
        {QGA-EE$_{T5}$ with 40\% training data} & {73.47} & {71.42} \\
    \hline
        {QGA-EE$_{T5}$ with 20\% training data} & {71.15} & {69.13} \\
    \hline
        
    \end{tabular}
    \caption{Performance of QGA-EE on ACE05 test data (F1, \%) with gold triggers with partial training data. Training data is randomly sampled.}
    \label{tab:data_aug}
\end{table}

%% file: 4.4_analysis.tex
\subsection{Analysis and Discussion}
\begin{figure}[h]
    \centering
    \includegraphics[width=0.9\linewidth]{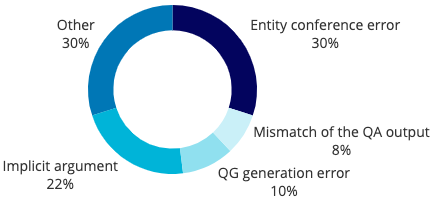}
    \caption{The portion of each category of error based on 50 error examples in test set.}
    \label{fig:error_analysis}
\end{figure}

The average length of the questions generated by QGA-EE$_{T5}$ is 10.5 tokens, compared with 6.7 in~\citet{du2020event}. They contain more context. 
For example, QGA-EE generates \textit{`Who was attacked by mob in state?'} for the \texttt{Target} role in \textit{`At least three members of a family in Indias northeastern state of Tripura were \textbf{[hacked$_{Conflict.Attack}$]} to death by a tribal mob for allegedly practicing witchcraft, police said Thursday.'} It incorporates  \texttt{Attacker} (`mob') and \texttt{Place} (`state') information.


We categorize the errors into four groups:
\begin{enumerate}
    \item Bad question generated by the QG model. 

    For example, QGA-EE generates \textit{`What did state buy in * sell * event?'} for the \texttt{Artifact} role in \textit{`... that the Stalinist state had developed nuclear weapons and hinted it may sell or use them, depending on US actions.'}. It should have been \textit{`What did state sell in * sell * event?'} and this introduces an error to the QA model.

     \item Errors resulting from a mismatch of the QA output result. QGA-EE may retrieve wrong offsets if a target candidate matches with multiple text strings in the original sentence. For example, QGA-EE matches the candidate \textit{`Welch'} with the first mention in \textit{`He also wants to subpoena all documents maintained in Jane Beasley \underline{Welch}'s personnel file by Shearman; Sterling, a prestigious corporate law firm where she worked before she \textbf{[married$_{Life.Marry}$]} \underline{Welch}.'}, where the correct one is the second mention.

     \item Errors resulting from missing entity conference. For example, QGA-EE identifies \textit{`Jacques Chirac'} as the \texttt{Entity} of the \texttt{Contact.Phone-Write} event in \textit{`French President Jacques Chirac received only a reserved response when he tried to mend fences by placing a telephone call Tuesday to Bush.'}. But \textit{`he'} is the ground truth and refers to \textit{`Jacques Chirac'}.

     \item Predictions not explicitly mentioned. For example, in \textit{`Kelly, the US assistant secretary for East Asia and Pacific Affairs, arrived in Seoul from Beijing Friday to brief Yoon, the foreign minister.'}, QGA-EE infers \textit{`Seoul'} as the \texttt{Place} of the \texttt{Contact.Meet} event, but it is not explicitly mentioned in the context, thus not covered by the gold annotations.
\end{enumerate}

We manually analyzed a subset of the errors from the test set (50 examples), and show the portion of each category of error in Figure~\ref{fig:error_analysis}.

%% file: 5_conclusion.tex
\section{Conclusion}
In this paper, we present QGA-EE, a novel sequence-to-sequence based framework for EE, which utilizes a QG model to generate context-aware questions as inputs to a QA model for EAE. Our model naturally supports the cases in which multiple event arguments play the same role within a specific event mention. We conduct experiments on the ACE05-E dataset and the proposed model outperforms all of the single-task-based models and achieves competitive results with state-of-the-art joint IE models.
In the future, we plan to utilize the extensibility of the QA framework to incorporate knowledge from semi-structured event-relevant data such as Wikipedia Infoboxes. We also plan to extend our approach to multilingual EE and joint IE.

%% file: 6_limitations.tex
\section*{Limitations}
The design of the dynamic templates requires knowledge of the event ontology and is time-consuming. The authors of the paper spent  30 hours designing the exclusive templates that cover all of the possible argument combinations for each argument role in ACE ontology. With a more complicated ontology, a much larger amount of time is required.

Another limitation of our approach is the offset retrieval method. If one sentence contains multiple mentions of the same entities, or even multiple text strings that have the same spellings but refer to different entities, the QGA-EE model always retrieves the position where the mention appears for the first time in the sentence as the offset of the extracted target. It may be improved by asking the model to generate contextual text as a position reference.

%% file: 8_Acknowledgements.tex
\section*{Acknowledgements}
We thank our colleague Aoife Cahill and the anonymous reviewers for their constructive comments and suggestions.

%% file: appex_b.tex
\section{ACE05-E Data Preprocessing}
\label{sec:appendix}
We follow the preprocessing steps in~\citet{wadden2019entity} to preprocess ACE2005 corpora. More specifically, we use the preprocessing script at \url{https://github.com/dwadden/dygiepp}. In addition, we retrieve the character positions of the event triggers and arguments, because T5 uses a SentencePiece tokenizer. Table~\ref{tab:data} shows the statistics of the ACE05-E dataset.

 \begin{table}[ht]
    \centering
    \begin{tabular}{c|c|c|c}
    \hline
    {Split}&{\#Sents}&{\#Events}&{\#Arguments}  \\
    \hline
    {Train}& {17,172}&{4,202}&{4,859} \\
    {Dev}& {923}&{450}&{605} \\
    {Test}& {832}&{403}&{576} \\
    \hline
    \end{tabular}
    \caption{Data statistics of the ACE05-E dataset.}
    \label{tab:data}
\end{table}

\section{Complete Dynamic Templates for ACE ontology}
Table~\ref{tab:templates_all} shows the complete list of templates with different combinations of known argument roles for each ACE event argument role.

\section{Implementation Details}
\label{app:implement}
We use Huggingface Transformers library~\cite{wolf-etal-2020-transformers} to load the model checkpoints.
\subsection{Event Trigger Labeling Model}
\begin{table}[!ht]
\centering
\begin{tabular}{cc}
\hline
 Hyperparamter & Value \\
\hline
Learning rate               & 3e-5           \\ 
Learning rate decay         & 1e-5           \\ 
Epoch                       & 20             \\ 
Batch size                  & 4              \\ 
Gradient accumulation steps & 4             \\ 
\hline
\end{tabular}
\caption{Hyperparameter for Event Trigger Labeling Model training.}
\label{tab:training_hp_trigger}
\end{table}
We implemented an ALBERT-based sequence labeling model for event trigger detection. We simply apply Softmax on top of the ALBERT encoder to predict the BIO schema based event label. We finetune the \texttt{albert-xxlarge-v2} checkpoint provided by Huggingface during training.~\footnote{\url{https://huggingface.co/albert-xxlarge-v2}}. We use the Adam optimizer with clip threshold of 1.0 and warmup proportion of 0.1.
Table~\ref{tab:training_hp_trigger} shows the hyperparameter to train the Event Trigger Labeling Model.

\subsection{QG model}
\begin{table}[!ht]
\centering
\begin{tabular}{cc}
\hline
 Hyperparamter & Value \\
\hline
Learning rate               & 1e-4           \\ 
Learning rate decay         & 1e-5           \\ 
Epoch                       & 20             \\ 
Batch size                  & 2              \\ 
Gradient accumulation steps & 32             \\ 
Number of beam & 4             \\
Length penalty & 0.0 \\
\hline
\end{tabular}
\caption{Hyperparameter for QG Model training.}
\label{tab:training_hp_qg}
\end{table}

When generating the groundtruth for QG model training, we use the basic template (e.g. `Who was the attacking agent?') without incorporating any arguments if the target event role does not exist in the event mention. And we do not restrict the QG model to generate verbs that only appear in the templates. They are preserved for training the QA model.

We finetune the \texttt{T5-large} checkpoint provided by Huggingface during training.~\footnote{\url{https://huggingface.co/t5-large}} with the Adafactor optimizer with clip threshold of 1.0 and warmup proportion of 0.1.
Table~\ref{tab:training_hp_qg} shows the hyperparameter to train the QG Model.
And Table~\ref{tab:training_statistics} shows the numbers of examples used to train and evaluate the QG model.
 \begin{table}[ht]
    \centering
    \begin{tabular}{c|c|c|c}
    \hline
    {}&{Train}&{Dev}&{Test}  \\
    \hline
    {QG Model}& {15,785}&{1,767}&{1,434} \\
    {QA Model}& {20,681}&{1,713}&{1,391} \\
    \hline
    \end{tabular}
    \caption{Number of examples used to train and evaluate the QG and QA models.}
    \label{tab:training_statistics}
\end{table}

\subsection{QA model}
\begin{table}[!ht]
\centering
\begin{tabular}{cc}
\hline
 Hyperparamter & Value \\
\hline
Learning rate               & 2e-4           \\ 
Learning rate decay         & 1e-5           \\ 
Epoch                       & 20             \\ 
Batch size                  & 2              \\ 
Gradient accumulation steps & 32             \\ 
Number of beam & 4             \\
Length penalty & -2.5 \\
\hline
\end{tabular}
\caption{Hyperparameter for QA Model training.}
\label{tab:training_hp_qa}
\end{table}

For the QA model training, we use the Adafactor optimizer with a learning rate of 2e-4, and weight decay of 1e-5, and clip threshold as 1.0. We set all of the relative\_step, scale\_parameter, and warmup\_init parameters to \texttt{False}. For optimizer scheduler, we set the warmup proportion to 0.1.

If there are no event arguments for the argument role, the output is empty, as the following example. We include them to train the QA model.
Table~\ref{tab:training_statistics} shows the numbers of examples used to train and evaluate the QA model.

\noindent\fbox{
\parbox{0.95\linewidth}{
\textbf{Input:} \textit{question: What device was used to inflict the harm in * injured * event? context: \underline{Injured} Russian diplomats and a convoy of America's Kurdish comrades in arms were among unintended victims caught in crossfire and friendly fire Sunday.}

 \textbf{Output:} $<$ /s $>$
 }
 }

In postprocessing, we dynamically change the start position for searching to keep the order of the retrieved event arguments.



\section{Experiment Details}
For all of the scores reported in the paper, the numbers are based on a single run with a fixed random seed 42. 

\subsection{Event Trigger Labeling Model}
Table~\ref{tab:trigger_score} shows the performance of the Event Trigger Labeling model on ACE05-E test set.
\begin{table}[ht]
    \centering
\begin{tabular}{ccc|ccc}
    \hline
    \multicolumn{3}{c|}{Trigger Identification} & \multicolumn{3}{c}{Trigger Classification}\\
    \hline
         P & R & F1 & P & R & F1  \\
         \hline
         {72.52} & {79.9} & {76.03} & {69.59} & {76.67} & {72.96}\\
         \hline
    \end{tabular}
    \caption{Performance of our event trigger labeling model on ACE05-E test data (\%).}
    \label{tab:trigger_score}
\end{table}

\subsection{QG Model}

We use Rouge~\cite{lin2004rouge} score (ROUGE-1) as the evaluation metric for QG model training, and the score on ACE05-E test set is 0.892.

\begin{table*}[]
    \centering
    \footnotesize
    \begin{tabular}{p{2cm}|p{1.6cm}|p{3cm}|p{6.5cm}}
        \hline
        \multirow{4}{*}{\texttt{Life.Be-Born}} & \multirow{2}{*}{\texttt{Person}} & - & Who was born? \\
        {} & {} & {\texttt{Place}} & {Who was born in [\texttt{Place}]?} \\
        \cline{2-4}
        {} & \multirow{2}{*}{\texttt{Place}} & - & Where did the birth take place? \\
        {} & {} & {\texttt{Person}} & {Where was [\texttt{Person}] born?} \\

        \hline
        \multirow{4}{*}{\texttt{Life.Marry}} & \multirow{2}{*}{\texttt{Person}} & - & Who was married? \\
        {} & {} & {\texttt{Place}} & {Who was married in [\texttt{Place}]?} \\
        \cline{2-4}
        {} & \multirow{2}{*}{\texttt{Place}} & - & Where did the marriage take place? \\
        {} & {} & {\texttt{Person}} & {Where was [\texttt{Person}] married?} \\
\end{tabular}
    \label{tab:templates_all}
\end{table*}

\begin{table*}[]
    \centering
    \footnotesize
    \begin{tabular}{p{2cm}|p{1.6cm}|p{3cm}|p{6.5cm}}
    \cline{2-4}
{} &  \multirow{8}{*}{\texttt{Instrument}} & - & What device was used to kill? \\
{} & {} & {\texttt{Agent}} & {What device did [\texttt{Agent}] use to kill others?} \\
{} & {} & {\texttt{Victim}} & {What device was used to kill [\texttt{Victim}]?} \\
{} & {} & {\texttt{Place}} & {What device was used to kill others in [\texttt{Place}]?} \\
{} & {} & {\texttt{Agent}, \texttt{Victim}} & {What device did [\texttt{Agent}] use to kill [\texttt{Victim}]?} \\
{} & {} & {\texttt{Agent}, \texttt{Place}} & {What device did [\texttt{Agent}] use to kill others in [\texttt{Place}]?} \\
{} & {} & {\texttt{Victim}, \texttt{Place}} & {What device was used to kill [\texttt{Victim}] in [\texttt{Place}]?} \\
{} & {} & {\texttt{Agent}, \texttt{Victim}, \texttt{Place}} & {What device did [\texttt{Agent}] use to kill [\texttt{Victim}] in [\texttt{Place}]?} \\
\end{tabular}
    \label{tab:my_label}
\end{table*}

\begin{table*}[]
    \centering
    \footnotesize
    \begin{tabular}{p{2cm}|p{1.6cm}|p{3cm}|p{6.5cm}}
    \cline{2-4}
    \multirow{32}{*}{\texttt{Movement.}} & \multirow{16}{*}{\texttt{Vehicle}} & - & What vehicle was used for transporting? \\
\multirow{32}{*}{\texttt{Transport}} & {} & {\texttt{Agent}} & {What vehicle did [\texttt{Agent}] use for transporting?} \\
{} & {} & {\texttt{Artifact}} & {What vehicle was used for transporting [\texttt{Artifact}]?} \\
{} & {} & {\texttt{Origin}} & {What vehicle was used for transporting from [\texttt{Origin}]?} \\
{} & {} & {\texttt{Destination}} & {What vehicle was used for transporting to [\texttt{Destination}]?} \\
{} & {} & {\texttt{Agent}, \texttt{Artifact}} & {What vehicle did [\texttt{Agent}] use for transporting [\texttt{Artifact}]?} \\
{} & {} & {\texttt{Agent}, \texttt{Origin}} & {What vehicle did [\texttt{Agent}] use for transporting from [\texttt{Origin}]?} \\
{} & {} & {\texttt{Agent}, \texttt{Destination}} & {What vehicle did [\texttt{Agent}] use for transporting to [\texttt{Destination}]?} \\
\end{tabular}
    \label{tab:my_label}
\end{table*}

\begin{table*}[]
    \centering
    \footnotesize
    \begin{tabular}{p{2cm}|p{1.6cm}|p{3cm}|p{6.5cm}}
    \multirow{8}{*}{} & \multirow{6}{*}{\texttt{Destination}} & {\texttt{Artifact}, \texttt{Vehicle}} & {Where was [\texttt{Artifact}] transported to with [\texttt{Vehicle}]?} \\
{} & {} & {\texttt{Artifact}, \texttt{Origin}} & {Where was [\texttt{Artifact}] transported to from [\texttt{Origin}]?} \\
{} & {} & {\texttt{Vehicle}, \texttt{Origin}} & {Where was artifact transported to from [\texttt{Origin}] with [\texttt{Vehicle}]?} \\
{} & {} & {\texttt{Agent}, \texttt{Artifact}, \texttt{Vehicle}} & {Where did [\texttt{Agent}] transport [\texttt{Artifact}] to with [\texttt{Vehicle}]?} \\
{} & {} & {\texttt{Agent}, \texttt{Artifact}, \texttt{Origin}} & {Where did [\texttt{Agent}] transport [\texttt{Artifact}] to from [\texttt{Origin}]?} \\
{} & {} & {\texttt{Agent}, \texttt{Vehicle}, \texttt{Origin}} & {Where did [\texttt{Agent}] transport artifact to from [\texttt{Origin}] with [\texttt{Vehicle}]?} \\
{} & {} & {\texttt{Artifact}, \texttt{Vehicle}, \texttt{Origin}} & {Where was [\texttt{Artifact}] transported to from [\texttt{Origin}] with [\texttt{Vehicle}]?} \\
{} & {} & {\texttt{Agent}, \texttt{Artifact}, \texttt{Vehicle}, \texttt{Origin}} & {Where did [\texttt{Agent}] transport [\texttt{Artifact}] to from [\texttt{Origin}] with [\texttt{Vehicle}]?} \\
\end{tabular}
    \label{tab:my_label}
\end{table*}

\begin{table*}[]
    \centering
    \footnotesize
    \begin{tabular}{p{2cm}|p{1.6cm}|p{3cm}|p{6.5cm}}
    \cline{2-4}
    \multirow{48}{*}{} & \multirow{16}{*}{\texttt{Beneficiary}} & - & Who benefits from the transaction? \\
{} & {} & {\texttt{Buyer}} & {Who did [\texttt{Buyer}] buy things for?} \\
{} & {} & {\texttt{Seller}} & {Who did buyer buy things from [\texttt{Seller}] for?} \\
{} & {} & {\texttt{Artifact}} & {Who did buyer buy [\texttt{Artifact}] for?} \\
{} & {} & {\texttt{Place}} & {Who did buyer buy things for in [\texttt{Place}]?} \\
{} & {} & {\texttt{Buyer}, \texttt{Seller}} & {Who did [\texttt{Buyer}] buy things from [\texttt{Seller}] for?} \\
{} & {} & {\texttt{Buyer}, \texttt{Artifact}} & {Who did [\texttt{Buyer}] buy [\texttt{Artifact}] for?} \\
{} & {} & {\texttt{Buyer}, \texttt{Place}} & {Who did [\texttt{Buyer}] buy things for in [\texttt{Place}]?} \\
\hline

    \end{tabular}
    \label{tab:my_label}

\end{table*}

\begin{table*}[]
    \centering
    \footnotesize
    \begin{tabular}{p{2cm}|p{1.6cm}|p{3cm}|p{6.5cm}}
    \hline
\multirow{32}{*}{\parbox{3.5cm}{\texttt{Transaction. Transfer-Money}}} & \multirow{8}{*}{\texttt{Giver}} & - & Who gave money to others? \\
{} & {} & {\texttt{Recipient}} & {Who gave money to [\texttt{Recipient}]?} \\
{} & {} & {\texttt{Beneficiary}} & {Who gave money to others for [\texttt{Beneficiary}]?} \\
{} & {} & {\texttt{Place}} & {Who gave money to others in [\texttt{Place}]?} \\
{} & {} & {\texttt{Recipient}, \texttt{Beneficiary}} & {Who gave money to [\texttt{Recipient}] for [\texttt{Beneficiary}]?} \\
{} & {} & {\texttt{Recipient}, \texttt{Place}} & {Who gave money to [\texttt{Recipient}] in [\texttt{Place}]?} \\
{} & {} & {\texttt{Beneficiary}, \texttt{Place}} & {Who gave money to others for [\texttt{Beneficiary}] in [\texttt{Place}]?} \\
{} & {} & {\texttt{Recipient}, \texttt{Beneficiary}, \texttt{Place}} & {Who gave money to [\texttt{Recipient}] for [\texttt{Beneficiary}] in [\texttt{Place}]?} \\
\hline

    \end{tabular}
    \label{tab:my_label}
\end{table*}

\begin{table*}[]
    \centering
    \footnotesize
    \begin{tabular}{p{2cm}|p{1.6cm}|p{3cm}|p{6.5cm}}
    \hline
\multirow{32}{*}{\texttt{Conflict.}} & \multirow{8}{*}{\texttt{Attacker}} & - & Who was the attacking agent? \\
\multirow{32}{*}{\texttt{Attack}} & {} & {\texttt{Target}} & {Who attacked [\texttt{Target}]?} \\
{} & {} & {\texttt{Instrument}} & {Who used [\texttt{Instrument}] in the attack?} \\
{} & {} & {\texttt{Place}} & {Who made the attack in [\texttt{Place}]?} \\
{} & {} & {\texttt{Target}, \texttt{Instrument}} & {Who attacked [\texttt{Target}] using [\texttt{Instrument}]?} \\
{} & {} & {\texttt{Target}, \texttt{Place}} & {Who attacked [\texttt{Target}] in [\texttt{Place}]?} \\
{} & {} & {\texttt{Instrument}, \texttt{Place}} & {Who used [\texttt{Instrument}] in the attack in [\texttt{Place}]?} \\
{} & {} & {\texttt{Target}, \texttt{Instrument}, \texttt{Place}} & {Who attacked [\texttt{Target}] using [\texttt{Instrument}] in [\texttt{Place}]?} \\
\hline
    \end{tabular}
    \label{tab:my_label}
\end{table*}

\begin{table*}[]
    \centering
    \footnotesize
    \begin{tabular}{p{2cm}|p{1.6cm}|p{3cm}|p{6.5cm}}
    \hline
\multirow{12}{*}{\parbox{3.5cm}{\texttt{Personnel. End-Position}}} & \multirow{4}{*}{\texttt{Person}} & - & Who ended the position? \\
{} & {} & {\texttt{Entity}} & {Who was fired by [\texttt{Entity}]?} \\
{} & {} & {\texttt{Place}} & {Who ended the position in [\texttt{Place}]?} \\
{} & {} & {\texttt{Entity}, \texttt{Place}} & {Who was fired by [\texttt{Entity}] in [\texttt{Place}]?} \\
\cline{2-4}
{} & \multirow{4}{*}{\texttt{Entity}} & - & Who fired employee? \\
{} & {} & {\texttt{Person}} & {Who fired [\texttt{Person}]?} \\
{} & {} & {\texttt{Place}} & {Who fired employee in [\texttt{Place}]?} \\
{} & {} & {\texttt{Person}, \texttt{Place}} & {Who fired [\texttt{Person}] in [\texttt{Place}]?} \\
\end{tabular}
    \label{tab:my_label}
\end{table*}

\begin{table*}[]
    \centering
    \footnotesize
    \begin{tabular}{p{2cm}|p{1.6cm}|p{3cm}|p{6.5cm}}
    \cline{2-4}
    \multirow{24}{*}{\parbox{2.5cm}{\texttt{Justice. Trial-Hearing}}} & \multirow{8}{*}{\texttt{Prosecutor}} & - & Who tried defendant? \\
{} & {} & {\texttt{Defendant}} & {Who tried [\texttt{Defendant}]?} \\
{} & {} & {\texttt{Adjudicator}} & {Who tried the defendant being adjudicated by [\texttt{Adjudicator}]?} \\
{} & {} & {\texttt{Place}} & {Who tried defendant in [\texttt{Place}]?} \\
{} & {} & {\texttt{Defendant}, \texttt{Adjudicator}} & {Who tried [\texttt{Defendant}] being adjudicated by [\texttt{Adjudicator}]?} \\
{} & {} & {\texttt{Defendant}, \texttt{Place}} & {Who tried [\texttt{Defendant}] in [\texttt{Place}]?} \\
{} & {} & {\texttt{Adjudicator}, \texttt{Place}} & {Who tried the defendant being adjudicated by [\texttt{Adjudicator}] in [\texttt{Place}]?} \\
{} & {} & {\texttt{Defendant}, \texttt{Adjudicator}, \texttt{Place}} & {Who tried [\texttt{Defendant}] being adjudicated by [\texttt{Adjudicator}] in [\texttt{Place}]?} \\
\end{tabular}
    \label{tab:my_label}
\end{table*}

\begin{table*}[]
    \centering
    \footnotesize
    \begin{tabular}{p{2cm}|p{1.6cm}|p{3cm}|p{6.5cm}}
    \cline{2-4}
\multirow{16}{*}{\parbox{2.5cm}{\texttt{Justice. Charge-Indict}}} & \multirow{8}{*}{\texttt{Adjudicator}} & - & Who adjudicated the indictment? \\
{} & {} & {\texttt{Defendant}} & {Who adjudicated the indictment [\texttt{Defendant}] was charged in?} \\
{} & {} & {\texttt{Prosecutor}} & {Who adjudicated the indictment executed by [\texttt{Prosecutor}]?} \\
{} & {} & {\texttt{Place}} & {Who adjudicated the indictment in [\texttt{Place}]?} \\
{} & {} & {\texttt{Defendant}, \texttt{Prosecutor}} & {Who adjudicated the indictment [\texttt{Defendant}] was charged in by [\texttt{Prosecutor}]?} \\
{} & {} & {\texttt{Defendant}, \texttt{Place}} & {Who adjudicated the indictment [\texttt{Defendant}] was charged in in [\texttt{Place}]?} \\
{} & {} & {\texttt{Prosecutor}, \texttt{Place}} & {Who adjudicated the indictment executed by [\texttt{Prosecutor}] in [\texttt{Place}]?} \\
{} & {} & {\texttt{Defendant}, \texttt{Prosecutor}, \texttt{Place}} & {Who adjudicated the indictment [\texttt{Defendant}] was charged in by [\texttt{Prosecutor}] in [\texttt{Place}]?} \\
\hline

    \end{tabular}
    \label{tab:my_label}
\end{table*}

\begin{table*}[]
    \centering
    \footnotesize
    \begin{tabular}{p{2cm}|p{1.6cm}|p{3cm}|p{6.5cm}}

    {} & \multirow{3}{*}{\texttt{Place}}  & {\texttt{Plaintiff}, \texttt{Defendant}} & {Where did [\texttt{Plaintiff}] sue [\texttt{Defendant}]?} \\
{} & {} & {\texttt{Plaintiff}, \texttt{Adjudicator}} & {Where did [\texttt{Plaintiff}] sue defendant being adjudicated by [\texttt{Adjudicator}]?} \\
{} & {} & {\texttt{Defendant}, \texttt{Adjudicator}} & {Where was [\texttt{Defendant}] sued being adjudicated by [\texttt{Adjudicator}]?} \\
{} & {} & {\texttt{Plaintiff}, \texttt{Defendant}, \texttt{Adjudicator}} & {Where did [\texttt{Plaintiff}] sue [\texttt{Defendant}] being adjudicated by [\texttt{Adjudicator}]?} \\
\hline
\multirow{12}{*}{\parbox{2.5cm}{\texttt{Justice. Convict}}} & \multirow{4}{*}{\texttt{Defendant}} & - & Who was convicted for crime? \\
{} & {} & {\texttt{Adjudicator}} & {Who was convicted by [\texttt{Adjudicator}] for crime?} \\
{} & {} & {\texttt{Place}} & {Who was convicted for crime in [\texttt{Place}]?} \\
{} & {} & {\texttt{Adjudicator}, \texttt{Place}} & {Who was convicted by [\texttt{Adjudicator}] for crime in [\texttt{Place}]?} \\
\end{tabular}
    \label{tab:my_label}
\end{table*}

\begin{table*}[]
    \centering
    \footnotesize
    \begin{tabular}{p{2cm}|p{1.6cm}|p{3cm}|p{6.5cm}}
    \cline{2-4}
    \multirow{6}{*}{\parbox{2.5cm}{\texttt{Justice. Extradite}}} & \multirow{4}{*}{\texttt{Destination}} & - & Where was the person extradited to? \\
{} & {} & {\texttt{Agent}} & {Where did [\texttt{Agent}] extradite person to?} \\
{} & {} & {\texttt{Origin}} & {Where was the person extradited to from [\texttt{Origin}]?} \\
{} & {} & {\texttt{Agent}, \texttt{Origin}} & {Where did [\texttt{Agent}] extradite person to from [\texttt{Origin}]?} \\
\cline{2-4}
{} & \multirow{4}{*}{\texttt{Origin}} & - & Where was the person extradited from? \\
{} & {} & {\texttt{Agent}} & {Where did [\texttt{Agent}] extradite person from?} \\
{} & {} & {\texttt{Destination}} & {Where was the person extradited from to [\texttt{Destination}]?} \\
{} & {} & {\texttt{Agent}, \texttt{Destination}} & {Where did [\texttt{Agent}] extradite person from to [\texttt{Destination}]?} \\
\hline
    \end{tabular}
    \caption{Complete Templates for argument roles in ACE ontology.}
    \label{tab:templates_all}
\end{table*}